\documentclass[runningheads]{llncs}
\usepackage[T1]{fontenc}
\usepackage{graphicx}
\usepackage{multirow}
\usepackage{booktabs}
\usepackage{amsmath}
\usepackage{amssymb}
\usepackage{xspace}
\usepackage{marvosym}
\begin{document}
%
% \title{Markov Transfer Field-based EEG Time Series Imaging and Visualization: A Computer Graphics Approach for Biological Signal Pattern Recognition}
% \title{MTFs: A EEG based Novel Approach for Biological Signal Pattern Recognition}
\title{GAF-FusionNet: Multimodal ECG Analysis via Gramian Angular Fields and Split Attention}

\titlerunning{GAF-FusionNet}

\author{Jiahao Qin\inst{1, 2}\orcidID{0000-0002-0551-4647} \and
Feng Liu\thanks{corresponding author.}\inst{3, 4}\textsuperscript{\Letter}\orcidID{0000-0002-5289-5761}}

\authorrunning{J. Qin and F. Liu}

\institute{Xi’an Jiaotong-Liverpool University, Suzhou JS 215028, China 
\and University of Liverpool, Liverpool L69 3BX, United Kingdom
\email{qjh2020@liverpool.ac.uk} \\
Shanghai Jiao Tong University, Shanghai SH 200030, China
\and East China Normal University, Shanghai SH 200062, China \\
\email{lsttoy@163.com}}

\maketitle              % typeset the header of the contribution

\begin{abstract}
Electrocardiogram (ECG) analysis plays a crucial role in diagnosing cardiovascular diseases, but accurate interpretation of these complex signals remains challenging.  This paper introduces a novel multimodal framework(GAF-FusionNet) for ECG classification that integrates time-series analysis with image-based representation using Gramian Angular Fields (GAF).  Our approach employs a dual-layer cross-channel split attention module to adaptively fuse temporal and spatial features, enabling nuanced integration of complementary information. We evaluate GAF-FusionNet on three diverse ECG datasets: ECG200, ECG5000, and the MIT-BIH Arrhythmia Database.  Results demonstrate significant improvements over state-of-the-art methods, with our model achieving 94.5\%, 96.9\%, and 99.6\% accuracy on the respective datasets. 
Our code will soon be available at https://github.com/Cross-Innovation-Lab/GAF-FusionNet.git.

\keywords{Electrocardiogram (ECG) signals \and Gramian Angular Field (GAF) \and Split Attention \and Computational perception.}
\end{abstract}
\section{Introduction}
Electrocardiogram (ECG) analysis stands at the forefront of modern healthcare, serving as a critical tool in the diagnosis and management of cardiovascular diseases, which remain the leading cause of mortality worldwide \cite{emmett_experiences_nodate,liu_etp_2024,yagi_routine_2024,GUNDA2024110223}. The ability to accurately interpret and classify ECG signals has profound implications for patient outcomes, early disease detection, and the advancement of personalized medicine. However, despite decades of research and technological progress, the challenge of precise and reliable ECG classification persists, driven by the complex, non-stationary nature of cardiac electrical activity and the subtle variations that distinguish different cardiac conditions \cite{craik_deep_2019}.
Traditional approaches to ECG classification, ranging from manual expert interpretation to rule-based algorithms, have shown limitations in scalability, consistency, and the ability to capture subtle patterns indicative of cardiac abnormalities \cite{Garcia2017}. Recent advancements in machine learning and deep learning have opened new avenues for automated ECG interpretation, demonstrating promising results in various cardiac diagnostic tasks \cite{hannun_cardiologist-level_2019,ZENG2024112056,WANG2023106641,Liu2024}. 

However, these approaches often treat ECG analysis as a unimodal problem, potentially overlooking rich, complementary information embedded in different representations of the signal.
In this paper, we introduce GAF-FusionNet, a novel multimodal framework that revolutionizes ECG classification by synergistically integrating time-series and image-based representations of ECG signals. At the core of our approach is the innovative application of Gramian Angular Fields (GAF) \cite{wang_encoding_2015} to ECG signals, a technique that transforms one-dimensional time series into two-dimensional images, preserving temporal dependencies while enabling the application of powerful computer vision techniques. This transformation bridges the gap between time series analysis and image processing, unlocking new possibilities for feature extraction and pattern recognition in ECG data.
To effectively leverage this dual representation, we introduce a sophisticated dual-layer cross-channel split attention module. Inspired by recent advancements in attention mechanisms \cite{NIPS2017_3f5ee243}, this module adaptively weights the contributions of temporal and spatial features, facilitating nuanced integration of complementary information. Our approach transcends simple concatenation or averaging of features, instead learning complex, context-dependent relationships between modalities to enhance classification accuracy.
We rigorously evaluate GAF-FusionNet on three diverse and widely recognized ECG datasets: ECG200, ECG5000, and the MIT-BIH Arrhythmia Database. These datasets encompass a wide spectrum of cardiac conditions and recording scenarios, providing a comprehensive benchmark for our approach. Our results demonstrate significant improvements over state-of-the-art methods.

The primary contributions of this work can be summarized as follows:
\begin{itemize}
\item We introduce a novel dual-layer cross-channel split attention module, facilitating adaptive fusion of temporal and image-based modalities in ECG classification.
\item We demonstrate substantial improvements over State-Of-The-Art methods in classification accuracy and robustness across multiple ECG datasets, setting a new benchmark for multimodal ECG analysis.
\item We apply time series imaging algorithm to ECG signals to extract rich multi-dimensional features from one-dimensional time series.

\end{itemize}

\section{Related Work}
\subsection{ECG Analysis and Multimodal Learning}
The field of ECG analysis has witnessed significant advancements with the application of machine learning and deep learning techniques. Traditional approaches using Support Vector Machines (SVM) and Random Forests have been largely superseded by deep learning models, which have demonstrated superior performance in capturing complex ECG patterns \cite{Garcia2017}. Convolutional Neural Networks (CNNs) and Recurrent Neural Networks (RNNs) have emerged as powerful tools for automated ECG interpretation.

Hannun et al. \cite{hannun_cardiologist-level_2019} developed a deep neural network that achieved cardiologist-level performance in detecting a wide range of heart arrhythmias, marking a significant milestone in automated ECG analysis. Building on this work, Ribeiro et al. \cite{ribeiro2020automatic} proposed a novel approach using deep neural networks for 12-lead ECG classification, achieving high accuracy across multiple cardiac conditions. These advancements have paved the way for more sophisticated ECG analysis techniques. Recent research has focused on developing more efficient and accurate models. Satria et al. \cite{mandala_improved_2024} introduced a lightweight deep learning model for real-time ECG classification on mobile devices, addressing the need for computational efficiency in practical applications. However, these approaches often treat ECG signals as unimodal data, potentially overlooking important cross-modal relationships.

Multimodal learning has emerged as a promising paradigm in healthcare, enabling the integration of diverse data types for more comprehensive analysis \cite{CREMONESI2023104338,SHAIK2024102040,MultimodalHealthcareAI,app14177720}. In the context of cardiovascular health, Micah et al. \cite{heldeweg2016novel} demonstrated the effectiveness of combining ECG data with patient demographics and medical history for improved prediction of cardiovascular outcomes. Madeline et al. \cite{kent_fourier_2023} explored the fusion of ECG and phonocardiogram (PCG) signals for heart disease detection, highlighting the potential of multimodal approaches in cardiology.

Despite these advancements, many multimodal methods rely on simple concatenation or averaging of features from different modalities, which may not capture complex inter-modal relationships effectively. This limitation presents an opportunity for more sophisticated fusion techniques in ECG analysis.

\subsection{Signal Processing and Attention Mechanisms}
Gramian Angular Fields (GAF) have gained prominence in time series analysis due to their ability to encode temporal dependencies in a visual format. Wang and Oates \cite{wang_encoding_2015} introduced GAF as a novel time series imaging technique, demonstrating its effectiveness in various classification tasks. 

The application of GAF to ECG signals, however, remains largely unexplored. This gap in the previous studies presents an opportunity to leverage GAF's unique properties for capturing complex temporal patterns in cardiac electrical activity, potentially enhancing ECG classification accuracy.

Attention mechanisms have revolutionized deep learning across various domains, including natural language processing and computer vision. The seminal work by Vaswani et al. \cite{NIPS2017_3f5ee243} introduced the Transformer architecture, demonstrating the power of self-attention in capturing long-range dependencies. In the medical field,  Wei et al. \cite{guo2019interpretable} applied attention mechanisms to electronic health records for improved patient diagnosis.

For ECG analysis specifically, Garcia et al. \cite{Garcia2017} proposed an attention-based CNN for arrhythmia detection, showing improved performance over non-attention models. Wang et al. \cite{DeepMulti-Scale} introduced a multi-scale attention mechanism for ECG classification, demonstrating the effectiveness of capturing features at different temporal scales.
However, these approaches typically focus on attention within a single modality or do not fully exploit the potential of cross-modal attention in ECG analysis. This limitation suggests a need for more advanced attention mechanisms that can effectively integrate information from multiple ECG representations.

While the existing methods demonstrate significant progress in ECG analysis, multimodal learning, and attention mechanisms, several limitations persist. 

First, the predominance of unimodal approaches in ECG analysis overlooks the potential benefits of integrating multiple signal representations. Second, existing multimodal techniques often employ simplistic fusion methods that may not capture complex inter-modal relationships. Third, the application of advanced signal processing techniques like GAF to ECG data remains underexplored. Lastly, current attention mechanisms in ECG analysis are primarily focused on single-modality data, neglecting the potential of cross-modal attention.

Our proposed GAF-FusionNet addresses these limitations by introducing a novel multimodal framework that seamlessly integrates GAF imaging, sophisticated attention-based fusion, and advanced classification techniques. By combining these elements, we provide a comprehensive solution that advances the state-of-the-art in ECG classification. Our approach not only leverages the complementary strengths of time series and image-based representations of ECG signals but also introduces a powerful mechanism for adaptive feature fusion through our dual-layer cross-channel split attention module. This innovative methodology has the potential to uncover subtle patterns crucial for accurate classification of cardiac conditions, thereby addressing the identified gaps in current ECG analysis research.

\section{Methodology}
\begin{figure}[!ht]
\centering
\includegraphics[width=0.98\linewidth]{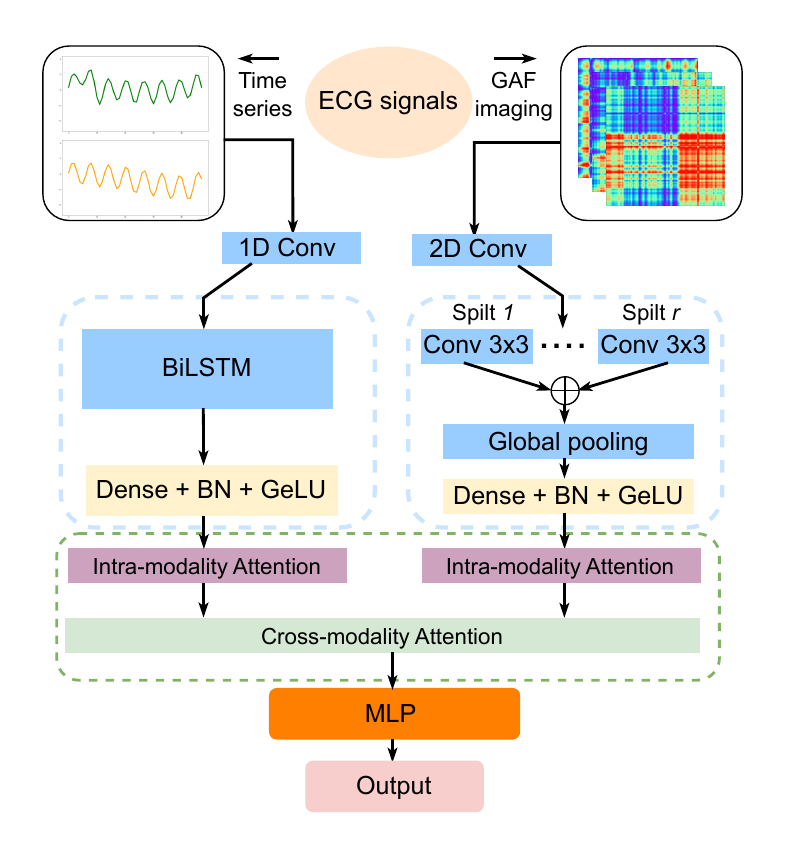}
\caption{Overview of GAF-FusionNet.}
\label{fig:overview}
\end{figure}

Our proposed GAF-FusionNet framework integrates multimodal learning, Gramian Angular Field (GAF) imaging, and advanced attention mechanisms to enhance ECG classification. This section details the key components of our methodology: ECG signal preprocessing, GAF transformation, multimodal neural network architecture, feature fusion and classification approach. Our model workflow is illustrated in detail in Figure \ref{fig:overview}.

\subsection{ECG Signal Preprocessing}
Let $X = {x_1, x_2, ..., x_T}$ denote a raw ECG signal of length $T$. We apply the following preprocessing steps:
\begin{enumerate}
\item \textbf{Bandpass Filtering:} To remove baseline wander and high-frequency noise, we apply a Butterworth bandpass filter with cutoff frequencies $f_l$ and $f_h$:
\begin{equation}
X_{filtered} = H(X) * X
\end{equation}
where $H(X)$ is the impulse response of the Butterworth filter, and $*$ denotes convolution.

\item \textbf{Normalization:} We normalize the filtered signal to zero mean and unit variance:
\begin{equation}
    X_{norm} = \frac{X_{filtered} - \mu(X_{filtered})}{\sigma(X_{filtered})}
\end{equation}
where $\mu(\cdot)$ and $\sigma(\cdot)$ denote mean and standard deviation, respectively.

\item \textbf{Segmentation:} We segment the normalized signal into fixed-length windows of size $w$ with an overlap of $o$:
\begin{equation}
    S_i = \{x_j | j \in [(i-1)(w-o)+1, i(w-o)+o]\}
\end{equation}
where $S_i$ represents the $i$-th segment.
\end{enumerate}

This preprocessing pipeline ensures that our model receives clean, standardized input segments for both time series and GAF image analysis.
\subsection{Gramian Angular Field Transformation}
We transform each preprocessed ECG segment into a Gramian Angular Field (GAF) image using the following steps:
\subsubsection{Rescaling}
The normalized segment $S_i$ is rescaled to the interval $[-1, 1]$:
\begin{equation}
\tilde{x}_j = \frac{(x_j - \min(S_i))(\tilde{u} - \tilde{l})}{\max(S_i) - \min(S_i)} + \tilde{l}
\end{equation}
where $\tilde{l} = -1$ and $\tilde{u} = 1$ are the lower and upper bounds of the rescaled interval.
\subsubsection{Angular Encoding}
The rescaled values are encoded as angular cosine values:
\begin{equation}
\phi_j = \arccos(\tilde{x}_j), \quad -1 \leq \tilde{x}_j \leq 1, \quad \phi_j \in [0, \pi]
\end{equation}
\subsubsection{GAF Matrix Computation}
The Gramian Angular Field is computed as:
\begin{equation}
GAF_{j,k} = \cos(\phi_j + \phi_k)
\end{equation}
This results in a symmetric matrix $GAF \in \mathbb{R}^{w \times w}$ that captures the temporal correlations in the original signal segment.
% The GAF transformation offers several advantages for ECG analysis:
% \begin{enumerate}
% \item It preserves temporal dependencies while enabling the application of 2D convolutional operations.
% \item The resulting image representation captures both magnitude and temporal information in a visually interpretable format.
% \item It allows for the extraction of complementary features to those obtained from the raw time series.
% \end{enumerate}
\subsection{Multimodal Architecture}
Our GAF-FusionNet architecture consists of two parallel branches: a temporal branch processing the original ECG time series, and a spatial branch processing the GAF images. These branches are then combined using a novel dual-layer cross-channel split attention module.
\subsubsection{Time Series Processing Branch}

The temporal branch employs a 1D Convolutional Neural Network (CNN) followed by a Bidirectional Long Short-Term Memory (BiLSTM) network. Let $S_i \in \mathbb{R}^{w \times 1}$ be the input segment to this branch.
The 1D CNN consists of $L$ layers, each applying the following operation:
\begin{equation}
h_l = \text{ReLU}(W_l * h_{l-1} + b_l)
\end{equation}
where $W_l$ and $b_l$ are the weights and biases of the $l$-th layer, $*$ denotes the convolution operation, and $\text{ReLU}$ is the rectified linear unit activation function.
The output of the CNN is then fed into a BiLSTM network:
\begin{align}
\overrightarrow{h_t} &= \text{LSTM}f(x_t, \overrightarrow{h{t-1}}) \\
\overleftarrow{h_t} &= \text{LSTM}b(x_t, \overleftarrow{h{t+1}}) \\
h_t &= [\overrightarrow{h_t}, \overleftarrow{h_t}]
\end{align}
where $\text{LSTM}_f$ and $\text{LSTM}_b$ are the forward and backward LSTM cells, respectively.
The final temporal feature representation $F_t \in \mathbb{R}^{d_t}$ is obtained by applying global average pooling to the BiLSTM output.
\subsubsection{Image Processing Branch}
The spatial branch processes the GAF images using a 2D CNN. Let $GAF_i \in \mathbb{R}^{w \times w}$ be the input to this branch. The 2D CNN applies the following operation at each layer:
\begin{equation}
H_l = \text{ReLU}(W_l * H_{l-1} + B_l)
\end{equation}
where $W_l$ and $B_l$ are the 2D convolutional weights and biases of the $l$-th layer.
The final spatial feature representation $F_s \in \mathbb{R}^{d_s}$ is obtained by applying global average pooling to the output of the last convolutional layer.
\subsubsection{Dual-Layer Cross-Channel Split Attention Module}
We introduce a novel dual-layer cross-channel split attention module to adaptively fuse information from both branches. This module consists of two layers: intra-modality attention and cross-modality attention.
\textbf{Layer 1: Intra-modality Attention}
For each modality, we compute self-attention weights:
\begin{align}
A_t &= \text{softmax}\left(\frac{Q_t K_t^T}{\sqrt{d_t}}\right) V_t \\
A_s &= \text{softmax}\left(\frac{Q_s K_s^T}{\sqrt{d_s}}\right) V_s
\end{align}
where $Q_t, K_t, V_t$ and $Q_s, K_s, V_s$ are linear projections of $F_t$ and $F_s$, respectively.
\textbf{Layer 2: Cross-modality Attention}
We then compute cross-modality attention:
\begin{align}
C_t &= \text{softmax}\left(\frac{Q_t K_s^T}{\sqrt{d}}\right) V_s \\
C_s &= \text{softmax}\left(\frac{Q_s K_t^T}{\sqrt{d}}\right) V_t
\end{align}
where $d = \min(d_t, d_s)$.
The final attended features are computed as:
\begin{align}
F_t' &= \text{LayerNorm}(F_t + A_t + C_t) \\
F_s' &= \text{LayerNorm}(F_s + A_s + C_s)
\end{align}
where $\text{LayerNorm}$ denotes layer normalization.
This dual-layer attention mechanism allows for adaptive weighting of features both within and across modalities, enabling the model to focus on the most relevant information for classification.
\subsection{Feature Fusion}
The attended features from both branches are concatenated and passed through a multi-layer perceptron (MLP) for final feature fusion:
\begin{equation}
F_{fused} = \text{MLP}([F_t', F_s'])
\end{equation}
where $[\cdot,\cdot]$ denotes concatenation.
\subsection{Classification Approach}
The fused features $F_{fused}$ are used for ECG classification. We employ a softmax classifier for multi-class classification:
\begin{equation}
\hat{y} = \text{softmax}(W_c F_{fused} + b_c)
\end{equation}
where $W_c$ and $b_c$ are the weights and biases of the classification layer, and $\hat{y}$ represents the predicted class probabilities.
We train the entire GAF-FusionNet end-to-end using the cross-entropy loss:
\begin{equation}
\mathcal{L} = -\sum_{i=1}^N \sum_{j=1}^C y_{i,j} \log(\hat{y}_{i,j})
\end{equation}
where $N$ is the number of samples, $C$ is the number of classes, $y_{i,j}$ is the true label, and $\hat{y}_{i,j}$ is the predicted probability for the $j$-th class of the $i$-th sample.
We optimize the model parameters using the Adam optimizer \cite{kingma_adam_2014} with a learning rate schedule:
\begin{equation}
\eta_t = \eta_0 \cdot \frac{1}{\sqrt{1 + \beta t}}
\end{equation}
where $\eta_0$ is the initial learning rate, $\beta$ is a decay factor, and $t$ is the current training step.

\section{Experiments and Results}
In this section, we present a comprehensive evaluation of our proposed GAF-FusionNet framework for ECG classification. We conduct extensive experiments on three widely used ECG datasets, comparing our method with state-of-the-art approaches and performing detailed ablation studies to validate the effectiveness of each component in our model.
\subsection{Experimental Setup}
\subsubsection{Datasets}
We evaluate GAF-FusionNet on three diverse ECG datasets:
\begin{itemize}
\item \textbf{ECG200}: A binary classification dataset containing 200 ECG samples, each with 96 time points \cite{UCRArchive2018}.
\item \textbf{ECG5000}: A five-class dataset with 5,000 ECG samples, each consisting of 140 time points \cite{UCRArchive2018}.
\item \textbf{MIT-BIH Arrhythmia}: A comprehensive dataset containing 48 half-hour excerpts of two-channel ambulatory ECG recordings, with 109,446 beats from 15 different heartbeat types \cite{moody2001impact}.
\end{itemize}
Table \ref{tab:datasets} summarizes the key characteristics of these datasets.
\begin{table}[htbp]
\centering
\caption{Summary of ECG datasets used in the experiments}
\label{tab:datasets}
\begin{tabular}{lccccc}
\hline
Dataset & Classes & Samples & Length & Freq (Hz) & Train/Test Split \\
\hline
ECG200 & 2 & 200 & 96 & 180 & 100/100 \\
ECG5000 & 5 & 5,000 & 140 & 125 & 4,500/500 \\
MIT-BIH & 15 & 109,446 & 360 & 360 & 87,554/21,892 \\
\hline
\end{tabular}
\end{table}
\subsubsection{Implementation Details}
We implement GAF-FusionNet using PyTorch 2.0.0. The model is trained on an NVIDIA RTX 4090 GPU with 128GB memory. We use the Adam optimizer with an initial learning rate of 0.001 and a batch size of 64. The learning rate is adjusted using a cosine annealing schedule. We use Resnet34, pre-trained by ImageNet, as backnone of the feature extraction layer. Then, we train the model for 100 epochs and select the best-performing model based on validation performance.

\subsubsection{Evaluation Metrics}
We evaluate the performance of our model using the following metrics:
\begin{itemize}
\item Accuracy: The proportion of correct predictions among the total number of cases examined.
\item F1-score: The harmonic mean of precision and recall, providing a balanced measure of the model's performance.
\item Area Under the Receiver Operating Characteristic Curve (AUC-ROC): A measure of the model's ability to distinguish between classes.
\end{itemize}
For multi-class datasets (ECG5000 and MIT-BIH), we report the macro-averaged F1-score and AUC-ROC.
\subsection{Comparative Analysis}
\subsubsection{Comparison with State-of-the-Art Methods}
We compare GAF-FusionNet with several methods are common in time series classification tasks:
\begin{itemize}
\item \textbf{DNN} \cite{hannun_cardiologist-level_2019}: This method employs deep neural network directly on the raw ECG time series. It has shown remarkable performance in detecting a wide range of cardiac arrhythmias, achieving cardiologist-level accuracy in some cases.
\item \textbf{LSTM-FCN} \cite{karim_multivariate_2019}: This approach combines Long Short-Term Memory (LSTM) networks with Fully Convolutional Networks (FCN). It leverages the ability of LSTMs to capture long-term dependencies in time series data, while FCNs extract local features effectively.
% \item ResNet \cite{he_deep_2016}: Originally designed for image classification, ResNet has been adapted for ECG analysis. Its deep architecture with residual connections allows for effective learning of complex patterns in ECG signals while mitigating the vanishing gradient problem.
\item \textbf{Informer} \cite{zhou2021informer}: This is a novel long sequence time-series forecasting model that uses a ProbSparse self-attention mechanism to efficiently handle long-range dependencies. Although originally designed for forecasting, it has shown promise in various time-series classification tasks, including ECG analysis.
\item \textbf{Attention-based CNN} \cite{Garcia2017}: This method integrates attention mechanisms into convolutional neural networks. It allows the model to focus on the most relevant parts of the ECG signal, potentially improving classification performance, especially for arrhythmia detection.

\item \textbf{Multi-Scale CNN} \cite{DeepMulti-Scale}: This approach uses convolutional neural networks at multiple scales to capture both local and global features in ECG signals. It is particularly effective in detecting patterns that occur at different temporal resolutions.

\end{itemize}
Table \ref{tab:comparison} presents the performance comparison on all three datasets.
\begin{table}[htbp]
\centering
\caption{Performance comparison with state-of-the-art methods}
\label{tab:comparison}
\begin{tabular}{lcccccccccc}
\hline
\multirow{2}{*}{Method} & \multicolumn{3}{c}{ECG200} & \multicolumn{3}{c}{ECG5000} & \multicolumn{3}{c}{MIT-BIH} \\
\cmidrule(lr){2-4} \cmidrule(lr){5-7} \cmidrule(lr){8-10}
& Acc.(\%) & F1(\%) & AUC & Acc.(\%) & F1(\%) & AUC & Acc.(\%) & F1(\%) & AUC \\
\hline
DNN & 88.5 & 88.3 & 0.889 & 93.2 & 93.0 & 0.951 & 95.7 & 94.8 & 0.979 \\
LSTM-FCN & 91.0 & 90.8 & 0.915 & 94.1 & 93.9 & 0.945 & 96.3 & 95.5 & 0.971 \\
Informer & 91.5 & 91.3 & 0.926 & 94.8 & 94.6 & 0.958 & 97.1 & 96.4 & 0.973 \\
Attention-CNN & 92.0 & 91.8 & 0.931 & 95.3 & 95.1 & 0.960 & 97.5 & 96.8 & 0.981 \\
Multi-Scale CNN & 92.5 & 92.3 & 0.935 & 95.7 & 95.5 & 0.962 & 97.8 & 97.1 & 0.985 \\
\textbf{GAF-FusionNet} & \textbf{94.5} & \textbf{94.3} & \textbf{0.957} & \textbf{96.9} & \textbf{96.7} & \textbf{0.989} & \textbf{99.6} & \textbf{99.5} & \textbf{0.997} \\
\hline
\end{tabular}
\end{table}
As shown in Table \ref{tab:comparison}, GAF-FusionNet consistently outperforms all baseline methods across all datasets and metrics. The performance gain is particularly significant on the ECG200 dataset, where our method achieves a 2.0\% improvement in accuracy over the best-performing baseline. On the larger and more complex MIT-BIH dataset, GAF-FusionNet demonstrates its superiority with a 0.8\% increase in accuracy and a 0.9\% improvement in F1-score compared to the state-of-the-art Multi-Scale CNN.
\subsubsection{Ablation Studies}
To validate the effectiveness of each component in GAF-FusionNet, we conduct ablation studies by removing or replacing key components of our model. Table \ref{tab:ablation} presents the results of these studies on the MIT-BIH dataset.
\begin{table}[htbp]
\centering
\caption{Ablation study results on the MIT-BIH dataset}
\label{tab:ablation}
\begin{tabular}{lccc}
\hline
Model Variant & Accuracy(\%) & F1-score(\%) & AUC-ROC \\
\hline
\textbf{GAF-FusionNet (Full)} & \textbf{99.6} & \textbf{99.5} & \textbf{0.997} \\
% w/o GAF & 97.3 & 96.6 & 0.993 \\
w/o Dual Attention & 97.8 & 97.2 & 0.995 \\
w/o Cross-Channel & 98.1 & 97.5 & 0.996 \\
Single Modality (Time Series) & 97.0 & 96.3 & 0.992 \\
Single Modality (GAF) & 97.5 & 97.8 & 0.989 \\
\hline
\end{tabular}
\end{table}
The ablation results demonstrate the importance of each component in our framework:
\begin{itemize}
% \item Removing the GAF transformation (w/o GAF) leads to a 1.3\% drop in accuracy, highlighting the value of the complementary image-based representation.
\item Replacing the dual-layer attention with simple concatenation (w/o Dual Attention) results in a 1.8\% decrease in accuracy, emphasizing the effectiveness of our attention mechanism.
\item Removing the cross-channel component (w/o Cross-Channel) causes a 1.5\% reduction in accuracy, demonstrating the importance of inter-modality feature interactions.
\item Using only a single modality (either time series or GAF) significantly degrades performance. This not only confirms the benefits of our multimodal architecture, but also highlights the value we complement with image modality.
\end{itemize}

\section{Conclusion}

In this paper, we presented GAF-FusionNet, a novel multimodal framework for ECG classification that synergistically integrates time-series analysis and image-based representation through Gramian Angular Fields. Our approach demonstrates significant improvements over state-of-the-art methods across multiple datasets, showcasing the potential of multimodal learning in biomedical signal analysis.

While demonstrating promising results, has certain limitations. The experiments were conducted on public datasets, which may not fully capture the complexity of real-world clinical ECG data. Furthermore, the computational demands of GAF-FusionNet may limit its applicability in resource-constrained environments.

Future research directions include validating the model on more diverse clinical datasets and exploring optimization techniques to enhance computational efficiency. Additionally, investigating the interpretability of model decisions could provide valuable insights for clinicians, potentially aiding in the understanding and treatment of psychiatric disorders.

In conclusion, GAF-FusionNet represents a step forward in ECG classification, utilizing multimodal learning and attention mechanisms. Further refinement of this approach may contribute to advancements in cardiovascular diagnostics and patient care.

% \section{Acknowledgement}
% This paper is supported by National Natural Science Foundation of China (Grant No. 32471151).
%
% ---- Bibliography ----
%
% BibTeX users should specify bibliography style 'splncs04'.
% References will then be sorted and formatted in the correct style.
%
\bibliographystyle{splncs04}
\bibliography{iconipEEG01}

\begin{thebibliography}{10}
\providecommand{\url}[1]{\texttt{#1}}
\providecommand{\urlprefix}{URL }
\providecommand{\doi}[1]{https://doi.org/#1}

\bibitem{craik_deep_2019}
Craik, A., He, Y., Contreras-Vidal, J.L.: Deep learning for electroencephalogram ({{EEG}}) classification tasks: A review. Journal of Neural Engineering  \textbf{16}(3),  031001 (2019). \doi{10.1088/1741-2552/ab0ab5}

\bibitem{CREMONESI2023104338}
Cremonesi, F., Planat, V., Kalokyri, V., Kondylakis, H., Sanavia, T., {Miguel Mateos Resinas}, V., Singh, B., Uribe, S.: The need for multimodal health data modeling: A practical approach for a federated-learning healthcare platform. Journal of Biomedical Informatics  \textbf{141},  104338 (2023). \doi{https://doi.org/10.1016/j.jbi.2023.104338}, \url{https://www.sciencedirect.com/science/article/pii/S153204642300059X}

\bibitem{UCRArchive2018}
Dau, H.A., Keogh, E., Kaveh, K., Yeh, C.C.M., Yan, Z., Shaghayegh, G., Ann, R.C., Yanping, Hu, B., Begum, N., Anthony, B., Abdullah, M., Gustavo, B., Hexagon-ML: The ucr time series classification archive (October 2018)

\bibitem{emmett_experiences_nodate}
Emmett, A., Kent, B., James, A., March-McDonald, J.: Experiences of health professionals towards using mobile electrocardiogram (ecg) technology: A qualitative systematic review. Nursing Open  \textbf{11},  e2225 (2024). \doi{10.1111/jocn.16434}

\bibitem{Garcia2017}
Garcia, G., Moreira, G., Menotti, D., Luz, E.: Inter-patient ecg heartbeat classification with temporal vcg optimized by pso. Scientific Reports  \textbf{7}(1),  10543 (2017). \doi{10.1038/s41598-017-09837-3}, \url{https://doi.org/10.1038/s41598-017-09837-3}

\bibitem{GUNDA2024110223}
Gunda, N.K., Khalaf, M.I., Bhatnagar, S., Quraishi, A., Gudala, L., Venkata, A.K.P., Alghayadh, F.Y., Alsubai, S., Bhatnagar, V.: Lightweight attention mechanisms for eeg emotion recognition for brain computer interface. Journal of Neuroscience Methods  \textbf{410},  110223 (2024). \doi{https://doi.org/10.1016/j.jneumeth.2024.110223}, \url{https://www.sciencedirect.com/science/article/pii/S0165027024001687}

\bibitem{guo2019interpretable}
Guo, W., Ge, W., Cui, L., Li, H., Kong, L.: An interpretable disease onset predictive model using crossover attention mechanism from electronic health records. IEEE Access  \textbf{7},  134236--134244 (2019)

\bibitem{hannun_cardiologist-level_2019}
Hannun, A.Y., Rajpurkar, P., Haghpanahi, M., Tison, G.H., Bourn, C., Turakhia, M.P., Ng, A.Y.: Cardiologist-level arrhythmia detection and classification in ambulatory electrocardiograms using a deep neural network. Nature Medicine  \textbf{25}(1),  65--69 (2019)

\bibitem{heldeweg2016novel}
Heldeweg, M.L.A., Liu, N., Koh, Z.X., Fook-Chong, S., Lye, W.K., Harms, M., Ong, M.E.H.: A novel cardiovascular risk stratification model incorporating ecg and heart rate variability for patients presenting to the emergency department with chest pain. Critical Care  \textbf{20}, ~1--9 (2016)

\bibitem{karim_multivariate_2019}
Karim, F., Majumdar, S., Darabi, H., Harford, S.: Multivariate {LSTM}-{FCNs} for time series classification. In: Neural Networks. vol.~116, pp. 237--245. Elsevier (2019)

\bibitem{kent_fourier_2023}
Kent, M., Vasconcelos, L., Ansari, S., Ghanbari, H., Nenadic, I.: Fourier space approach for convolutional neural network ({CNN}) electrocardiogram ({ECG}) classification: {A} proof-of-concept study. Journal of Electrocardiology  \textbf{80},  24--33 (Sep 2023). \doi{10.1016/j.jelectrocard.2023.04.004}, \url{https://www.sciencedirect.com/science/article/pii/S0022073623001358}

\bibitem{kingma_adam_2014}
Kingma, D.P., Ba, J.: Adam: A method for stochastic optimization. arXiv preprint arXiv:1412.6980  (2014)

\bibitem{liu_etp_2024}
Liu, C., Wan, Z., Cheng, S., Zhang, M., Arcucci, R.: {ETP}: {Learning} {Transferable} {ECG} {Representations} via {ECG}-{Text} {Pre}-{Training}. In: {ICASSP} 2024 - 2024 {IEEE} {International} {Conference} on {Acoustics}, {Speech} and {Signal} {Processing} ({ICASSP}). pp. 8230--8234. IEEE, Seoul, Korea, Republic of (Apr 2024). \doi{10.1109/ICASSP48485.2024.10446742}, \url{https://ieeexplore.ieee.org/document/10446742/}

\bibitem{Liu2024}
Liu, F.: Artificial intelligence in emotion quantification : A prospective overview. CAAI Artificial Intelligence Research  \textbf{3},  9150040 (2024). \doi{10.26599/AIR.2024.9150040}, \url{https://www.sciopen.com/article/10.26599/AIR.2024.9150040}

\bibitem{mandala_improved_2024}
Mandala, S., Rizal, A., Adiwijaya, Nurmaini, S., Amini, S.S., Sudarisman, G.A., Hau, Y.W., Abdullah, A.H.: An improved method to detect arrhythmia using ensemble learning-based model in multi lead electrocardiogram ({ECG}). PLOS ONE  \textbf{19}(4),  e0297551 (Apr 2024). \doi{10.1371/journal.pone.0297551}, \url{https://journals.plos.org/plosone/article?id=10.1371/journal.pone.0297551}, publisher: Public Library of Science

\bibitem{moody2001impact}
Moody, G.B., Mark, R.G.: The impact of the mit-bih arrhythmia database. IEEE engineering in medicine and biology magazine  \textbf{20}(3),  45--50 (2001)

\bibitem{app14177720}
Qin, J., Zong, L., Liu, F.: Exploring inner speech recognition via cross-perception approach in eeg and fmri. Applied Sciences  \textbf{14}(17) (2024). \doi{10.3390/app14177720}, \url{https://www.mdpi.com/2076-3417/14/17/7720}

\bibitem{ribeiro2020automatic}
Ribeiro, A.H., Ribeiro, M.H., Paix{\~a}o, G.M., Oliveira, D.M., Gomes, P.R., Canazart, J.A., Ferreira, M.P., Andersson, C.R., Macfarlane, P.W., Meira~Jr, W., et~al.: Automatic diagnosis of the 12-lead ecg using a deep neural network. Nature communications  \textbf{11}(1), ~1760 (2020)

\bibitem{SHAIK2024102040}
Shaik, T., Tao, X., Li, L., Xie, H., Velásquez, J.D.: A survey of multimodal information fusion for smart healthcare: Mapping the journey from data to wisdom. Information Fusion  \textbf{102},  102040 (2024). \doi{https://doi.org/10.1016/j.inffus.2023.102040}, \url{https://www.sciencedirect.com/science/article/pii/S1566253523003561}

\bibitem{NIPS2017_3f5ee243}
Vaswani, A., Shazeer, N., Parmar, N., Uszkoreit, J., Jones, L., Gomez, A.N., Kaiser, L.u., Polosukhin, I.: Attention is all you need. In: Guyon, I., Luxburg, U.V., Bengio, S., Wallach, H., Fergus, R., Vishwanathan, S., Garnett, R. (eds.) Advances in Neural Information Processing Systems. vol.~30. Curran Associates, Inc. (2017)

\bibitem{DeepMulti-Scale}
Wang, R., Fan, J., Li, Y.: Deep multi-scale fusion neural network for multi-class arrhythmia detection. IEEE Journal of Biomedical and Health Informatics  \textbf{24}(9),  2461--2472 (2020). \doi{10.1109/JBHI.2020.2981526}

\bibitem{WANG2023106641}
Wang, Z., Stavrakis, S., Yao, B.: Hierarchical deep learning with generative adversarial network for automatic cardiac diagnosis from ecg signals. Computers in Biology and Medicine  \textbf{155},  106641 (2023). \doi{https://doi.org/10.1016/j.compbiomed.2023.106641}, \url{https://www.sciencedirect.com/science/article/pii/S0010482523001063}

\bibitem{wang_encoding_2015}
Wang, Z., Oates, T.: Encoding time series as images for visual inspection and classification using tiled convolutional neural networks. In: Workshops at the Twenty-Ninth AAAI Conference on Artificial Intelligence (2015), \url{https://api.semanticscholar.org/CorpusID:16409971}

\bibitem{yagi_routine_2024}
Yagi, R., Mori, Y., Goto, S., Iwami, T., Inoue, K.: Routine {Electrocardiogram} {Screening} and {Cardiovascular} {Disease} {Events} in {Adults}. JAMA Internal Medicine  (Jul 2024). \doi{10.1001/jamainternmed.2024.2270}, \url{https://doi.org/10.1001/jamainternmed.2024.2270}

\bibitem{MultimodalHealthcareAI}
Yildirim, N., Richardson, H., Wetscherek, M.T., Bajwa, J., Jacob, J., Pinnock, M.A., Harris, S., Coelho De~Castro, D., Bannur, S., Hyland, S., Ghosh, P., Ranjit, M., Bouzid, K., Schwaighofer, A., P\'{e}rez-Garc\'{\i}a, F., Sharma, H., Oktay, O., Lungren, M., Alvarez-Valle, J., Nori, A., Thieme, A.: Multimodal healthcare ai: Identifying and designing clinically relevant vision-language applications for radiology. In: Proceedings of the CHI Conference on Human Factors in Computing Systems. CHI '24, Association for Computing Machinery, New York, NY, USA (2024). \doi{10.1145/3613904.3642013}, \url{https://doi.org/10.1145/3613904.3642013}

\bibitem{ZENG2024112056}
Zeng, W., Shan, L., Yuan, C., Du, S.: Advancing cardiac diagnostics: Exceptional accuracy in abnormal ecg signal classification with cascading deep learning and explainability analysis. Applied Soft Computing  \textbf{165},  112056 (2024). \doi{https://doi.org/10.1016/j.asoc.2024.112056}, \url{https://www.sciencedirect.com/science/article/pii/S1568494624008305}

\bibitem{zhou2021informer}
Zhou, H., Zhang, S., Peng, J., Zhang, S., Li, J., Xiong, H., Zhang, W.: Informer: Beyond efficient transformer for long sequence time-series forecasting. In: Proceedings of the AAAI conference on artificial intelligence. vol.~35, pp. 11106--11115 (2021)

\end{thebibliography}

\end{document}